\pdfoutput=1  

\PassOptionsToPackage{table,dvipsnames}{xcolor}

\documentclass{applemlr}
\usepackage{amsmath}
\usepackage{enumerate}
\usepackage{algorithm}
\usepackage{algpseudocode}
\usepackage{amsfonts}
\usepackage{amsthm}
\usepackage{cleveref}
\usepackage{diagbox}
\usepackage{colortbl}
\usepackage{amssymb}
\usepackage{xspace}
\usepackage{wrapfig}
\usepackage{adjustbox}
\usepackage{tabularx}
\usepackage{booktabs}
\usepackage{mathtools}
\usepackage{tikz}
\usepackage{enumitem}
\usepackage{silence}
\usepackage{dsfont}
\usepackage[table]{xcolor}
\usepackage[dvipsnames]{xcolor}
\usepackage{multirow}
\usepackage{makecell}
\usepackage{xfakebold}

\usepackage{amsmath,amsfonts,bm}

\def\eqref#1{equation~\ref{#1}}

\def\1{\bm{1}}

\DeclareMathAlphabet{\mathsfit}{\encodingdefault}{\sfdefault}{m}{sl}
\SetMathAlphabet{\mathsfit}{bold}{\encodingdefault}{\sfdefault}{bx}{n}

\definecolor{textgray}{HTML}{6E6E73}
\usetikzlibrary{positioning, calc}
\usetikzlibrary{decorations.pathmorphing}

\makeatletter
\patchcmd{\wrong@fontshape}{\@gobbletwo}{}{}{}
\makeatother
\numberwithin{equation}{section}
\makeatletter
\AtBeginDocument{
  \urlstyle{sf}
  
}
\makeatother

\definecolor{light}{RGB}{125, 125, 125}
\crefname{tcb@cnt@pbox}{code}{code}
\Crefname{tcb@cnt@pbox}{Code}{Code}
\crefname{assumption}{assumption}{assumption}
\Crefname{assumption}{Assumption}{Assumptions}

\newtcolorbox[auto counter]{pbox}[2][]{
  colback=white,
  title=Code~\thetcbcounter: #2,
  #1,fonttitle=\sffamily,
  fontupper=\sffamily,
  arc=2pt,
  colframe=bgcolor,
  coltitle=fgcolor,
  colbacktitle=bgcolor,
  toptitle=0.25cm,
  bottomtitle=0.125cm
}

\makeatletter
\newcommand\applefootnote[1]{%
  \begingroup
  \renewcommand\thefootnote{}%
  \renewcommand\@makefntext[1]{\noindent##1}%
  \footnote{#1}%
  \addtocounter{footnote}{-1}%
  \endgroup
}
\makeatother

\definecolor{cverbbg}{gray}{0.90}

\usepackage{textcomp}
\usepackage{array}

\graphicspath{{figures/}}

\newif\ifshownotes
\shownotesfalse       %

\ifshownotes
    
    \newcommand{\hintcite}[1]{\textcolor{red}{\textbf{[CITE: #1]}}}
    \newcommand{\mdsl}[1]{\textcolor{blue}{\textbf{[MDS: #1]}}}
\else
    
    \newcommand{\hintcite}[1]{}
    \newcommand{\mdsl}[1]{}
\fi

\definecolor{mds2pink}{RGB}{214,10,120}
\ifshownotes
    \newcommand{\MDSLII}[1]{\textcolor{mds2pink}{\textbf{[MDS2]}\ #1}}
\else
    \newcommand{\MDSLII}[1]{}
\fi

\definecolor{widecopy}{RGB}{0,110,110}

\makeatletter
\renewenvironment{table*}{\@float{table}}{\end@float}
\renewenvironment{figure*}{\@float{figure}}{\end@float}
\makeatother

\let\cite\citep

\title{Language Discrimination Improves Linguistic Learning in Multilingual Speech Models}

\author{Maureen de Seyssel}
\author[*]{Jie Chi}
\author[*]{Zakaria Aldeneh}
\contribution[*]{Equal contribution}

\affiliation{Apple}

\abstract{
Multilingual self-supervised speech models can benefit from sharing information across languages, but under a matched total pretraining data budget they still fall short of monolingual models. We show that strengthening the model's ability to discriminate languages during pretraining reduces and, on some measures, closes this multilingual gap on continuous phonetic and higher-level linguistic measures, while preserving substantial cross-language sharing. Using a controlled English/French HuBERT setting, we test two interventions which strengthen language discrimination: an auxiliary language classifier and per-language $k$-means targets. Across interventions, continuous-feature phone discrimination error (phone-ABX, $\downarrow$) decreases from $11.6\%$ in the bilingual baseline to $10.4\%$ (monolingual: $10.8\%$), while lexical performance (sWUGGY, $\uparrow$) increases from $52.1\%$ to $56.7\%$ (monolingual: $58.5\%$) and prosodic performance (ProsAudit, lexical subtask, $\uparrow$) from $68.9\%$ to $72.9\%$ (monolingual: $72.6\%$). Across HuBERT training stages, the strongest gains on most linguistic measures occur when language discrimination is introduced in the first iteration, whereas later or repeated interventions yield smaller improvements and are accompanied by increased language-wise segregation. These results support a causal role for language discrimination in reducing the additional cost of multilingual learning.

}

\metadata[Keywords]{\sffamily self-supervised learning, multilingual speech, representation learning, spoken language modelling}
\date{\sffamily\today}
\metadata[Correspondence]{\sffamily \url{mdeseyssel@apple.com}}

\begin{document}

\maketitle

\section{Introduction}

Self-supervised speech modelling has made it possible to learn linguistic structure directly from speech, with particular promise for textless NLP \cite{dunbar2022self,lakhotia2021generative}. So far, however, most work on learning linguistic structure from raw speech has focused on monolingual settings. Multilingual self-supervised speech models offer the possibility of learning several languages within a single encoder and sharing acoustic and linguistic structure across them \cite{conneau2021xlsr}. These shared representations can allow knowledge learned from one language to benefit learning in another, reducing the amount of language-specific data needed to learn shared structure. This should theoretically allow multilingual models to approach monolingual performance with less data per language, rather than requiring pretraining data to scale with the number of languages. Yet, at a matched total pretraining budget, multilingual encoders can still lag behind monolingual ones in linguistic learning, resulting in a multilingual gap \cite{deseyssel2023unsupervised,blandon2026leveraging} and suggesting an additional cost of multilingual learning.  The gap also persists after quantisation: spoken language models trained on quantised multilingual representations can lag behind monolingual ones on zero-shot lexical and prosodic evaluations \cite{deseyssel2023unsupervised,nguyen2020zerospeech,deseyssel2023prosaudit}. This raises the question of whether multilingual models can approach monolingual performance under reduced per-language exposure, and, if so, what currently prevents them from fully benefiting from multilingual training.

One possible source of this additional cost relates to how well multilingual models discriminate between languages. Human bilingual learners can acquire two languages despite reduced exposure to each, with language discrimination abilities present from birth \cite{bosch1997native,byers2010roots}. In contrast, multilingual self-supervised speech models do not develop robust language discrimination in their representations \cite{deseyssel2023unsupervised}. Previous work found that stronger language discrimination in multilingual speech models is associated with smaller multilingual gaps as training data increases \cite{deseyssel2023unsupervised} and under visual grounding \cite{blandon2026leveraging}, but these results do not establish causality. 
 One reason language discrimination may matter is that multilingual learning must balance transfer, where one language benefits from another, and interference, where information from one language can negatively affect learning in another.
Shared representations support transfer, while discriminating languages may help avoid treating acoustically similar material from different languages as equivalent. We therefore hypothesise that multilingual models benefit from discriminating languages sufficiently to reduce interference, as long as it preserves the shared representations needed for cross-language transfer.

In this work, we test whether language discrimination, as operationalised in previous work \cite{deseyssel2023unsupervised,blandon2026leveraging}, contributes causally to linguistic learning in multilingual speech models by applying two interventions designed to strengthen it in a controlled English/French bilingual setting. We adopt the matched-budget setup of \cite{deseyssel2023unsupervised} and apply it to HuBERT, holding the architecture, language pair, and total pretraining data budget fixed across conditions.
We show that \textbf{strengthening language discrimination during pretraining reduces the multilingual gap} on continuous phonetic and higher-level linguistic measures. In our HuBERT setting, the strongest gains occur when language discrimination is introduced in iteration one, whereas later or sustained interventions are accompanied by increased language-wise segregation and do not yield comparable improvements.

\section{Experimental Setup}
\label{sec:setup}

\subsection{Data and pretraining}
\label{sec:data}

We train English, French, and bilingual HuBERT models from scratch on read speech, using Libri-Light \cite{kahn2020libri} for English and Audiocite \cite{felice2024audiocite} for French. Our main comparison matches total pretraining data: the monolingual reference models use 1000~hours, while the bilingual model uses 500~hours per language. We also train 500~hours monolingual controls on the exact bilingual subsets and a bilingual model with 1000~hours per language to assess the effect of per-language exposure. The experimental conditions are summarised in Table~\ref{tab:levers}.

All models use 95M-parameter HuBERT-Base \cite{hsu2021hubert} and follow the standard two-iterations training procedure. Iteration one predicts $K{=}100$ MFCC $k$-means targets. For iteration two, we cluster iteration-one layer-6 representations with $K{=}500$ and retrain HuBERT from scratch on these targets. We then discretise iteration-two layer-9 representations with a shared $K{=}50$ $k$-means tokenizer and train a 3-layer LSTM spoken language model (SLM) on the encoder pretraining corpus, following the ZeroSpeech 2021 baselines \cite{nguyen2020zerospeech}. The SLM is used for the lexical and prosodic evaluations described below. Each condition is repeated over three full pretraining seeds, varying iteration-one training, target generation, iteration-two training, and SLM training. %

\subsection{Evaluation}
\label{sec:evaluation}

We evaluate both how language is represented in the models and how well the resulting representations support linguistic learning.

\subsubsection{Language information}

We use the same English and French Common Voice subsets as in STELA \cite{ardila2020common,lavechin_stela} to characterise how language is represented in the models. Our main measure is language discrimination: whether the model can tell one language apart from another. We also measure language separability and segregation : looking at these measures alongside discrimination helps clarify what we mean by language discrimination and distinguish it from other ways in which language information can be represented. Separability asks whether language identity can be recovered from the representation at all, whereas segregation asks whether representations are organised into language-specific neighbourhoods. This distinction matters because language can be easy to decode without being strongly discriminative or segregated.

\textbf{Language discrimination} is measured with across-speaker language ABX error between English and French ($\downarrow$) \cite{deseyssel2023unsupervised,blandon2026leveraging} on utterance-mean-pooled representations using angular distance. Chance is $50\%$, and we report the lowest error across encoder layers.
\textbf{Language separability} is measured with an $L_2$-regularised linear language probe ($\downarrow$) on utterance-mean representations, using held-out classification error. Chance is $50\%$, and we report the lowest error across encoder layers.
\textbf{Language segregation} is measured with the phone-segment-pooled soft nearest-neighbour same-language probability $P_{\mathrm{same}}$ \cite{frosst2019snnl}. It measures the distance-weighted probability that neighbouring points have the same language label. With balanced English and French data, values near $0.5$ indicate little language-wise segregation, while values approaching $1$ indicate increasingly language-pure neighbourhoods. We report the maximum $P_{\mathrm{same}}$ across encoder layers.

\subsubsection{Linguistic evaluation}

We also evaluate linguistic learning at the phonetic, lexical, and prosodic levels. 
\textbf{Phonetic learning} is measured with within-speaker phone ABX error ($\downarrow$) \cite{schatz2013evaluating}, following the STELA setup \cite{lavechin_stela}, on English and French Common Voice phone contrasts. We report two variants: \emph{continuous phone ABX}, computed on continuous encoder representations using angular distance, and \emph{unit phone ABX}, computed on the final $K{=}50$ layer-9 units used by the SLM. For continuous phone ABX, we report the lowest error across encoder layers.

The other linguistic evaluations are performed on the SLMs and assess higher-level linguistic learning, including lexical and prosodic structure. \textbf{Lexical learning} is evaluated using the English and French sWUGGY sets \cite{nguyen2020zerospeech,lavechin_stela}. \textbf{Prosodic learning} is evaluated with the English ProsAudit benchmark \cite{deseyssel2023prosaudit}, which includes lexical and protosyntactic subtasks.\footnote{To test whether the SLM results are specific to the language-modelling architecture, we also evaluated sWUGGY with a BERT-small masked unit LM. The same qualitative pattern across conditions was preserved.}

Following the convention used in STELA \cite{lavechin_stela}, we report English/French means for paired linguistic evaluations: monolingual English models are evaluated on English, monolingual French models on French, and bilingual models on both languages. ProsAudit is evaluated on English only as there is no French equivalent.

\vspace{4pt}
These experiments use public research datasets and standard HuBERT architectures in a controlled research setting and are not intended to describe a production speech system.

\section{Characterising the bilingual baseline}\label{sec:baseline}
We first examine the matched-budget bilingual baseline through two questions: how is language represented, and where does multilingual learning fall short?

\subsection{Discrimination, separability, and segregation}

In the matched-budget bilingual HuBERT, language identity is almost perfectly separable (linear-probe error $<1\%$), yet language discrimination is comparatively weak (lang-ABX $=27.0\%$) and segregation remains low ($P_{\mathrm{same}}\approx0.51$; Table~\ref{tab:levers}). 
The same dissociation is also observed across multilingual architectures, including XLSR-53 \cite{conneau2021xlsr}, w2v-BERT 2.0 \cite{barrault2023seamless}, and mHuBERT-147 \cite{boito2024mhubert}. On English and French, language identity remains highly separable, while lang-ABX stays between $24.9\%$ and $28.3\%$ and $P_{\mathrm{same}}$ remains close to $0.5$.\footnote{As a robustness check, $P_{\mathrm{same}}$ also remains low after utterance pooling, reaching at most $0.55$ across layers in the bilingual baseline.} These results show that weak language discrimination can coexist with near-perfect decodability and little language-wise segregation. Because stronger discrimination could also come from greater separation between languages, we track segregation throughout the interventions.

\begin{table*}[t]
\caption{\textbf{Language discrimination and linguistic evaluation across intervention and reference conditions.} Mean (SD) over three training seeds; phone ABX and sWUGGY report EN/FR means, with monolingual results averaged over English-to-English and French-to-French evaluations; ProsAudit is English-only.  \emph{Classifier} adds an auxiliary language classifier during iteration one; \emph{per-lang. targets} uses separate language-specific HuBERT target codebooks during iteration two; \emph{placebo (shuffled)} uses the classifier with shuffled language labels. Reference conditions are \emph{bilingual 2000h}, \emph{monolingual 1000 h}, and \emph{monolingual 500 h}.}

\label{tab:levers}
\centering
\small
\setlength{\tabcolsep}{3pt}

\adjustbox{max width=\linewidth}{%
\begin{tabular}{@{}l c c c c c c c c@{}}
\toprule
& & \multicolumn{2}{c}{language} &
\multicolumn{2}{c}{phone-ABX\,$\downarrow$} &
sWUGGY (\%)\,$\uparrow$ &
\multicolumn{2}{c}{ProsAudit (\%)\,$\uparrow$} \\
\cmidrule(lr){3-4}
\cmidrule(lr){5-6}
\cmidrule(lr){7-7}
\cmidrule(lr){8-9}

condition & pretraining data &
lang-ABX (\%)\,$\downarrow$ & $P_{\text{same}}$ &
cont. (\%) & $K{=}50$ (\%) &
& lexical & protosyntax \\
\midrule

baseline
& 500h EN + 500h FR
& 27.0 (4.4)
& 0.51 (0.002)
& 11.6 (0.9)
& 20.3 (0.4)
& 52.1 (2.1)
& 68.9 (3.5)
& 63.4 (1.3) \\

classifier
& 500h EN + 500h FR
& 5.2 (2.3)
& 0.52 (0.003)
& \textbf{10.4 (0.4)}
& \textbf{20.1} (1.2)
& 55.2 (3.2)
& \textbf{72.9 (1.3)}
& 67.4 (4.7) \\

per-lang. targets
& 500h EN + 500h FR
& \textbf{2.9} (1.0)
& 0.54 (0.01)
& 11.2 (0.1)
& 20.4 (1.0)
& \textbf{56.7} (1.5)
& 71.9 (3.3)
& \textbf{70.8 (7.2)} \\

\addlinespace[2pt]

\textit{placebo (shuffled)}
& 500h EN + 500h FR
& 24.2 (0.4)
& 0.51 (0.001)
& 11.1 (0.5)
& 20.8 (1.1)
& 51.5 (1.1)
& 68.3 (4.0)
& 67.8 (7.8) \\

\midrule

bilingual, 2000h
& 1000h EN + 1000h FR
& 29.0 (3.3)
& 0.51 (0.01)
& 11.1 (0.5)
& 20.8 (0.6)
& 53.2 (1.8)
& 69.7 (1.4)
& 66.7 (6.5) \\

mono, 1000 h
& 1000h mono
& n/a
& n/a
& 10.8 (0.5)
& 19.3 (0.5)
& 58.5 (2.1)
& 72.6 (0.7)
& 62.8 (6.6) \\

mono, 500 h
& 500h mono
& n/a
& n/a
& 11.2 (0.2)
& 19.0 (0.2)
& 56.5 (0.3)
& 72.3 (3.2)
& 66.0 (2.1) \\

\bottomrule
\end{tabular}
}%
\end{table*}

\subsection{The bilingual gap}
\label{sec:gap}

At a matched total pretraining budget, the bilingual model underperforms the monolingual references on continuous phone ABX ($11.6\%$ versus $10.8\%$), sWUGGY ($52.1\%$ versus $58.5\%$), and the lexical ProsAudit task ($68.9\%$ versus $72.6\%$; Table~\ref{tab:levers}). No bilingual gap is observed on ProsAudit protosyntax, consistent with previous work \cite{deseyssel2023unsupervised}. This may reflect the more cross-linguistically shared cues supporting this task. 
Finally, unit phone ABX is similar to the 1000 hours monolingual reference and only 1.3 points higher than the 500 hours monolingual control, despite the substantial lexical gap. This indicates that phone discrimination of the final discrete inventory alone does not explain lexical learning.

Reducing monolingual pretraining from 1000 to 500~hours has little effect (continuous phone ABX: $10.8\% \rightarrow 11.2\%$; sWUGGY: $58.5\% \rightarrow 56.5\%$), and the 500~hours monolingual models still outperform the bilingual model with the same per-language exposure. This shows that reduced per-language exposure alone does not explain the bilingual gap, supporting an additional cost of joint multilingual learning. Conversely, increasing bilingual pretraining to 1000~hours per language also does not fully recover monolingual performance.

\section{Manipulating Language Discrimination}
\label{sec:manipulations}

We then ask whether inducing stronger language discrimination reduces this gap, and whether the effect depends on how and at which stage discrimination is introduced.

\subsection{Comparison of intervention strategies}
\label{sec:intervention-results}

The baseline results show a bilingual gap that is not explained by reduced per-language exposure alone, while language discrimination remains weak. To test whether language discrimination contributes to this gap, we apply two interventions designed to strengthen it while tracking $P_{\mathrm{same}}$ to ensure that the languages do not simply become more segregated. Both affect iteration-two training, but introduce language information differently: the classifier explicitly supervises language identity during iteration one, whereas per-language targets introduce the distinction directly through the HuBERT prediction targets. Neither intervention requires language identity at test time. %

Our first intervention adds an auxiliary two-way language classifier at layer 6 during iteration one, which is also the layer used to construct iteration-two targets. Its cross-entropy loss is weighted by $\lambda_{\mathrm{LID}}=0.3$, ramped linearly over the first 32k steps, and added to the HuBERT objective. The classifier is used only during iteration one and discarded before iteration-two target generation. All evaluations are performed on the final iteration-two encoder.
We also train a shuffled-label placebo, randomly reassigning language labels within each batch, to control for effects of adding the auxiliary classifier independently of language-label information. The classifier substantially strengthens language discrimination (lang-ABX: 27.0\% $\rightarrow$ 5.2\%) while $P_{\mathrm{same}}$ remains low. Continuous phone ABX reaches 10.4\%, that is even slightly better than the monolingual reference (10.8\%), whereas unit phone ABX remains essentially unchanged. sWUGGY reaches 55.2\%, narrowing the gap to the monolingual reference (58.5\%). ProsAudit lexical performance reaches 72.9\%, slightly exceeding the monolingual reference (72.6\%), while protosyntactic performance also improves despite the absence of a baseline bilingual gap. The shuffled-label placebo has little to no effect on language discrimination and does not reproduce the classifier's gains on sWUGGY or lexical ProsAudit.

\begin{table*}[!t]

\caption{\textbf{Effect of auxiliary language classification at different HuBERT iterations.}
All models are trained on bilingual data (500\,h EN + 500\,h FR).
Mean (SD) over three training seeds; phone ABX and sWUGGY report EN/FR means; ProsAudit is English-only.
\emph{Classifier @iter1}, \emph{@iter2}, and \emph{@both} apply the auxiliary language classifier during iteration one, iteration two, or both, respectively.}
\label{tab:localize}
\centering
\small
\setlength{\tabcolsep}{4pt}

\adjustbox{max width=\linewidth}{%
\begin{tabular}{@{}l c c c c c c c@{}}
\toprule
& \multicolumn{2}{c}{language} &
\multicolumn{2}{c}{phone-ABX\,$\downarrow$} &
sWUGGY &
\multicolumn{2}{c}{ProsAudit} \\
\cmidrule(lr){2-3}
\cmidrule(lr){4-5}
\cmidrule(lr){6-6}
\cmidrule(lr){7-8}

condition &
ABX (\%) & $P_{\text{same}}$ &
cont. & $K{=}50$ &
(\%) & lex. & proto. \\
\midrule

baseline
& 27.0 (4.4)
& 0.51 (0.002)
& 11.6 (0.9)
& 20.3 (0.4)
& 52.1 (2.1)
& 68.9 (3.5)
& 63.4 (1.3) \\

classifier @iter1
& 5.2 (2.3)
& 0.52 (0.003)
& \textbf{10.4} (0.4)
& 20.1 (1.2)
& \textbf{55.2} (3.2)
& \textbf{72.9} (1.3)
& 67.4 (4.7) \\

classifier @iter2
& 2.8 (2.0)
& 0.55 (0.02)
& 11.8 (0.2)
& \textbf{19.3} (1.1)
& 53.0 (1.8)
& 66.2 (1.6)
& 65.8 (1.2) \\

classifier @both
& \textbf{2.5} (1.2)
& 0.57 (0.03)
& 11.9 (0.7)
& 19.8 (0.3)
& 53.6 (0.8)
& 67.0 (3.3)
& \textbf{69.1} (0.9) \\

\bottomrule
\end{tabular}
}%
\end{table*}

We next test \textbf{per-language HuBERT targets}, which strengthen language discrimination without an explicit language-classification objective. Iteration~1 is unchanged, but the shared $K{=}500$ iteration-two inventory is replaced by two language-specific $K{=}250$ codebooks with separate indices (i.e., representations from each language are quantized independently into 250 targets). This modifies the prediction targets without adding model capacity or increasing the per-language inventory.
Language discrimination again increases strongly (lang-ABX: $27.0\% \rightarrow 2.9\%$) while $P_{\mathrm{same}}$ remains low ($0.54$). Continuous phone ABX reaches $11.2\%$, narrowing the gap to the monolingual reference, while unit phone ABX remains comparable to baseline. Both sWUGGY and ProsAudit lexical performance narrow the gap to the monolingual references, with sWUGGY reaching $56.7\%$ and ProsAudit lexical $71.9\%$. ProsAudit protosyntactic performance also increases, despite the absence of a baseline gap. 

Across both interventions, phone discrimination improves on the continuous representations but not on the final $K{=}50$ units. The SLM gains therefore cannot be explained by better phone discrimination in the final discrete units, suggesting the gains are not simply due to a better phonetic tokenizer. More importantly, the interventions also reduce the multilingual gap relative to the 1000~hours monolingual references: the classifier matches or exceeds monolingual performance on continuous phone ABX ($10.4\%$ versus $10.8\%$) and ProsAudit lexical ($72.9\%$ versus $72.6\%$), while per-language targets substantially narrow the sWUGGY gap ($56.7\%$ versus $58.5\%$). These gains are obtained despite the bilingual models seeing only 500~hours of each language. Although the two interventions operate differently, both strengthen language discrimination while keeping $P_{\mathrm{same}}$ low. Together with the 500~hours monolingual controls, these results support a causal benefit of stronger language discrimination that cannot be explained by reduced per-language exposure alone.

\subsection{Intervention stage in HuBERT}
\label{sec:localize}

Both interventions above affect the targets used to train iteration-two. The auxiliary classifier changes the iteration-one representations that are subsequently clustered, while per-language HuBERT targets change the targets directly.
We therefore test how the stage at which language discrimination is introduced affects learning by applying the classifier at iteration-one, iteration-two, or both. The iteration-two condition acts only after target construction, whereas the both-iterations condition maintains the objective throughout pretraining.

All classifier schedules strongly increase language discrimination, but the strongest reductions in the linguistic gap are specific to the iteration-one intervention (Table~\ref{tab:localize}). Unit phone ABX is an exception, reaching its lowest value at iteration two, but this is not accompanied by comparable gains on the SLM metrics, suggesting that strong final language discrimination is not sufficient. The iteration-two and both-iterations conditions are also accompanied by greater segregation: $P_{\mathrm{same}}$ increases from $0.52$ at iteration one to $0.55$ at iteration two and $0.57$ when the classifier is applied at both iterations. 
This greater segregation may partly offset the benefit of discrimination by reducing cross-language sharing.

As a complementary control, we also apply language-specific clustering only to the final $K{=}50$ SLM tokenizer, using two disjoint $K{=}25$ codebooks. sWUGGY increases only slightly ($52.1\%$ to $53.0\%$), while ProsAudit lexical does not improve on any seed. Making the final tokenizer language-specific therefore does not reproduce the gains from introducing language discrimination during HuBERT pretraining.

\section{Discussion}

Across two interventions, strengthening language discrimination during pretraining improves linguistic learning while preserving substantial cross-language sharing, supporting a discriminable but shared regime. The interventions allow the bilingual model to approach the performance of the 1000~hours monolingual references on several measures despite seeing only 500~hours of each language, while the 500~hours monolingual controls show that this cannot be explained by reduced per-language exposure alone.
The gains on the SLM metrics also occur without improved phone discrimination in the final discrete units, suggesting that they are not simply due to better phonetic learning. This indicates that the benefits of language discrimination extend beyond improvements in the final phonetic representations, to higher-level linguistic learning.

The two interventions provide converging evidence about the role of language discrimination despite introducing it in different ways: the classifier explicitly supervises language identity, whereas per-language targets introduce the discrimination through the HuBERT prediction targets. Their common effect suggests that the relevant factor is language discrimination rather than a specific training objective or target construction.

The stage experiment further shows that the benefit of language discrimination is not determined by its final strength alone. Although the iteration-two and both-iteration interventions produce stronger final language discrimination, they do not yield comparable linguistic gains and are accompanied by greater language-wise segregation. This suggests that the timing of introducing language discrimination matters, and that stronger final discrimination alone is not necessarily beneficial when accompanied by greater language-wise segregation. Moreover, the final-tokenizer control shows that making the final discrete units language-specific does not reproduce the original gains.
This pattern is consistent with a transfer/interference account, in which sharing supports cross-lingual transfer while discrimination reduces interference, although we do not measure these mechanisms separately. Overall, our results suggest that multilingual speech models benefit from strengthening language discrimination where it can guide linguistic learning, while preserving the cross-language sharing needed to learn from reduced per-language exposure.

These findings should be interpreted within the controlled setting of our experiments. We deliberately adopt a tightly controlled setup to isolate the role of language discrimination, keeping the language pair, encoder architecture, and training data fixed while varying how the distinction is introduced. This supports a causal interpretation within our English/French HuBERT setting, but leaves open how broadly the findings generalise across architectures and languages: language pairs may differ in how readily they become discriminable and how much they benefit from stronger language discrimination. Our evidence for the transfer/interference framing is also indirect: we observe the predicted pattern of stronger discrimination with limited segregation and improved linguistic learning, rather than directly decomposing transfer and interference.
Finally, our localisation results depend on HuBERT's iterative target-generation procedure. Extending this principle to non-iterative models and to settings without explicit language labels remains important future work.

\section{Conclusion}
\label{sec:conclusion}

Overall, strengthening language discrimination during multilingual pretraining substantially reduces the additional cost of joint multilingual learning. \textbf{Despite seeing only 500~hours of each language, our best bilingual models close the continuous phonetic gap to the 1000~hours monolingual references and substantially reduce the gap on higher-level linguistic measures.} These gains do not occur with strong language-wise segregation: the useful regime appears to be discriminable but shared. The gains also extend beyond phonetic learning, with higher-level linguistic measures improving even when phone discrimination in the final discrete units does not. Together, the controlled interventions provide evidence that language discrimination during multilingual pretraining plays a causal role in these linguistic gains. Within HuBERT, however, strong language discrimination in the \emph{final} representations is not sufficient. This raises two questions: whether the main role of language discrimination is to influence the model during training rather than persist in the final representations, and how to strengthen language discrimination while preserving the shared regime that enables cross-language transfer.

\bibliographystyle{plainnat}
\bibliography{references}

\applefootnote{\textcolor{textgray}{\sffamily Apple and the Apple logo are trademarks of Apple Inc., registered in the U.S. and other countries and regions.}}

\end{document}